%% file: main.tex
\documentclass[sigconf]{acmart}

\usepackage{booktabs} % For formal tables

\setcitestyle{square}

\usepackage[ruled]{algorithm2e} % For algorithms

\SetAlFnt{\small}
\SetAlCapFnt{\small}
\SetAlCapNameFnt{\small}
\SetAlCapHSkip{0pt}
\IncMargin{-\parindent}

\acmJournal{TOG}
\acmVolume{9}
\acmNumber{4}
\acmArticle{39}
\acmYear{2010}
\acmMonth{3}

\setcopyright{usgovmixed}
\acmDOI{0000001.0000001_2}

\begin{document}
% Title portion
\title{Recent Progress in Conversational AI}

\author{Zijun Xue}
\orcid{1234-5678-9012-3456}
\affiliation{%
  \institution{UCLA}
  \streetaddress{104 Jamestown Rd}
  \city{Los Angeles}
  \state{CA}
  \postcode{90024}
  \country{USA}}
\email{xuezijun@cs.ucla.edu}
\author{Ruirui Li}
\affiliation{%
  \institution{UCLA}
  \streetaddress{104 Jamestown Rd}
  \city{Los Angeles}
  \state{CA}
  \postcode{90024}
  \country{USA}}
\email{rrli@cs.ucla.edu}
\author{Mingda Li}
\affiliation{%
  \institution{UCLA}
  \streetaddress{104 Jamestown Rd}
  \city{Los Angeles}
  \state{CA}
  \postcode{90024}
  \country{USA}}
\email{limingda@cs.ucla.edu}

\begin{abstract}
Conversational artificial intelligence (AI) is becoming an increasingly popular topic among industry and academia. With the fast development of neural network-based models, a lot of neural-based conversational AI system are developed. We will provide a brief review of the recent progress in the Conversational AI, including the commonly adopted techniques, notable works, famous competitions from academia and industry and widely used datasets.
\end{abstract}

\begin{CCSXML}
<ccs2012>
 <concept>
  <concept_id>10010520.10010553.10010562</concept_id>
  <concept_desc>Computer systems organization~Embedded systems</concept_desc>
  <concept_significance>500</concept_significance>
 </concept>
 <concept>
  <concept_id>10010520.10010575.10010755</concept_id>
  <concept_desc>Computer systems organization~Redundancy</concept_desc>
  <concept_significance>300</concept_significance>
 </concept>
 <concept>
  <concept_id>10010520.10010553.10010554</concept_id>
  <concept_desc>Computer systems organization~Robotics</concept_desc>
  <concept_significance>100</concept_significance>
 </concept>
 <concept>
  <concept_id>10003033.10003083.10003095</concept_id>
  <concept_desc>Networks~Network reliability</concept_desc>
  <concept_significance>100</concept_significance>
 </concept>
</ccs2012>
\end{CCSXML}

\ccsdesc[500]{Computer systems organization~Embedded systems}
\ccsdesc[300]{Computer systems organization~Redundancy}
\ccsdesc{Computer systems organization~Robotics}
\ccsdesc[100]{Networks~Network reliability}


\keywords{Conversational AI, Chatbot, Generative Model}

\maketitle
\input{introduction}

\input{review}

\input{works}

\input{datasets}

\bibliographystyle{ACM-Reference-Format}
\bibliography{sample-bibliography}
\end{document}

%% file: introduction.tex
\section{Introduction}
Conversational AI is a long-standing research topic. Both academia and industry organizations showed great interest for this type of systems. The conversational AI system has great commercial value and involves a lot of interesting questions ranging from natural language processing, speech recognition, knowledge base reasoning and human-computer interaction design etc. There are a number of large-scale conversational AI systems had been built, such as Siri, Xiaoice, Alexa and Google Assistant.

Recently, with the surge of neural-based models in various fields \cite{radfar2020end,li2019learning,vaswani2017attention,he2016deep,li2018deep,fu2020magnn} , a variety of neural-based conversational AI systems have been developed\cite{Haixun,octopus,alessandra2017roving,eigen,bowden2018slugbot,fang2017sounding}. The main techniques adopted by researchers are three categories: the distributed representation of entities, the sequence-to-sequence model and the reinforcement learning framework. The distributed representation is adopted to represent the internal status, user utterance, and external knowledge to enable more convenient retrieval and processing. The sequence-to-sequence model is applied to generate high quality general-purpose conversational AI systems. The reinforcement learning method is developed more and more in the task-specific or goal-oriented systems by optimizing the overall performance of the whole task.

Traditionally, the conversational chatbot consists of three parts: the natural language understanding unit, the dialogue manager, and the response generation unit. The natural language understanding unit \cite{rodd2017study,li2020multi,li2020efficient} is responsible for convert the best interpretation or n-best interpretations of user speech \cite{liu2021asr,li2020improving} generated from the automatic speech recognition (ASR) into the internal belief representation. The dialogue manager will be responsible for the process the internal representation and pick up a proper policy to generate a proper output. The response generation unit will be based on the dialogue manager's belief to generate a response for the user. It could be a text reply, an order for a hotel or a system API call etc.

The conversational AI could be classified as casual chat or chit-chat AI and task-oriented AI. For chit-chat AI, the purpose of the AI is conducting meaningful discussion as the daily casual talk between human beings. These types of AI usually doesn't need to refer to database or external information. For the task-oriented AI, the user assumes the AI can provide reliable task-specific information so that this type of AI usually requires to query an external database or knowledge base.

In this paper, we will mainly introduce the recent progress of conversational AI with more focus on the neural-based methods. 

We will introduce several widely applied neural network models in conversational AI in section 2. We will introduce the recent works in section 3. We will introduce several open conversational AI challenge and public datasets in section 4. We will draw brief concluding remarks in section 5.

%% file: review.tex
\section{Notable Models}
There are two prominent classes of models which are widely adopted by various conversational AI systems. The first class of model is the reinforcement learning model\cite{DBLP:journals/corr/MnihKSGAWR13}. This type of model is widely adopted in task-oriented dialogue models. The second category is the sequence-to-sequence model which is first introduced by Sutskever et al\cite{DBLP:journals/corr/SutskeverVL14}. The sequence-to-sequence model is widely adopted in general-purpose dialogue models.
\subsection{Reinforcement Learning Model}
DQN\cite{DBLP:journals/corr/MnihKSGAWR13} is a reinforcement learning framework which is proposed by Mnih et in 2013. This work is known for its success in plating Atari games without any predefined dataset. This work inspired numerous later work by incorporating this technique into different scenarios.

LeakGAN~\cite{guo2017long} is proposed to address the reward sparsity issue. It further uses features extracted from the discriminator as a step-by-step guide to train the generator. It improves the performance of sequence generation, especially in long text.

RankGAN~\cite{DBLP:conf/nips/LinLHSZ17} claims that the richness inside the sentences constrained by binary prediction is too restrictive and relaxes the training of the discriminator to a learning-to-rank optimization problem.

\subsection{Sequence-to-sequence Model}
the long short-term memory LSTM\cite{Hochreiter:1997:LSM:1246443.1246450}is a type of recurrent neural network which is designed to process the sequential data. The sequence-to-sequence model is widely applied to the conversational AI systems. 

The sequence-to-sequence model is proposed by Sutskever, Le et al~\cite{DBLP:journals/corr/SutskeverVL14}. This model utilizes the LSTM structure and designed an encoder-decoder structure to process the conversation data.

SeqGAN~\cite{yu2017seqgan} applies adversarial training on sequence generation under the reinforcement learning framework. 
SeqGAN models the text generation as a sequential decision-making task. The state is previously generated words, the action is the next word to be generated, and the rewards are the feedback derived from the discriminator. 
More precisely, the generator, constructed by an LSTM model, generates a sequence of words. The generated sequence of words is passed to the discriminator to generate reward signals.
The discriminator is constructed by a binary classifier. The binary classifier takes sentences from real data as positive examples and takes sentences generated by $G$ as negative examples.
The reward is defined by the cross-entropy between the predicted distribution and true distribution.

%% file: works.tex
\section{Recent works}
In this section, we will introduce a number of recent attempt in conversational AI system. We categorize these works into two parts: task-oriented system and general purpose conversational AI. The former one is usually domain-specific and potentially tend to consult an external database or knowledge bases. The latter one is the topics which are shared amount different types of conversational systems and they investigate various aspects which will affect the style and satisfactory of human users.
\subsection{RL-based method and Task-oriented system}
In recent research works, task-oriented dialog system has been modeled as a reinforcement learning system. The reward of the system is decided by whether the overall task is finished. Previous evaluation for generative system evaluates the quality of single generated sentence pair which doesn't take the overall task into consideration.

In \cite{young2013pomdp}, the author proposed a partially observed Markov decision process(POMDP) for a dialog process. The key point of this work is the state belief tracing and reinforcement learning. It consists of the language model M and the policy model P. The language model generates a belief state and policy model generates a response based on the internal parameter and the belief state from the language model. The accumulated results will be accumulated and evaluated by the reward function.

Gunasekara in \cite{gunasekaraquantized} tried to model the dialogue process by separating the language and logic. The author uses the cluster of entities to replace the original entity. A language model is trained based on this abstract representation.  The cluster is generated by conglomerate the nearby word embeddings. The dialog states are also tracked in order to achieve high relevance with the question.

In \cite{li2016deep}, Li proposed a method to incorporate the reinforcement learning for dialog generation. The challenge for adopting the reinforcement learning in dialog generation is the design of reward function. In this paper, the author proposed three aspects for the design of reward function: informativity (non-repetitive turns), coherence, and ease of answering (related to forward-looking function).

In \cite{wang2017integrating}, the author proposed a reinforcement seq2seq model, which is called a user and agent model integration framework(SAMIA) to generate the task-oriented dialog. This work is based on the observation that the asymmetric behavior of the role in the task-oriented dialog. One role, the user is trained by the seq2seq model and another role, the agent is trained by the reinforcement learning. They adopt a coffee ordering dataset to verify the effectiveness of this model.

In \cite{peng2017composite}, the author proposes a hierarchical deep reinforcement learning framework to solve the composite task-completion dialogue policy learning problem. The notable point of this paper is the problem is formulated as a Markov Decision Process and a hierarchical reinforcement learning method is used to train a dialogue manager. The dialogue manager consists of a top-level manager, a subgoal manager, and a global manager to ensure the overall goal is finished.

In \cite{ilievski2018goal}, the author tries to solve the lack of data problem for the task-specific problem by using a transfer learning method. The author claims that the lack of training data for goal-oriented systems are prevalent. The transfer learning method can improve the performance of distant tasks by 20\% and double the performance for close related tasks.

In \cite{joshi2017personalization,lipton2017bbq,toxtli2018understanding}, various other aspects and applications of goal-oriented dialogue system is investigated. In [52]\cite{joshi2017personalization} the author proposed a personalization in goal-oriented dialog. The author tries to model the personalization as a multi-task learning problem. They also suggest a single model which shares features among various profiles is better than separate models for each profile. In [53]\cite{lipton2017bbq}, the author proposed an acceleration technique by using the Thompson sampling to improve the efficiency of the deep reinforcement learning(DQN). In [54]\cite{toxtli2018understanding}, the author investigates how chatbot could be used to coordinate teamwork. The author deployed a prototype system to coordinate eight teams to finish create, assign, and keep track of the tasks. Seven insights for the future design of the chatbot are also proposed.

\subsection{end2end Task-oriented system}
For the end-to-end networks, there are two prominent categories. The first category\cite{sukhbaatar2015end,bordes2016learning,dodge2015evaluating} is mainly based on the Memory Network[memory]. The second category is attempting to access the external library and converting the input questions into some internal representations.

In \cite{wen2016network}, the author model the task-oriented problem as a sequence-to-sequence mapping problem which is augmented by the dialogue history and external information. The sequence-to-sequence modeling can generate a proper format of the response with some empty "slots". The input will be processed by two internal systems. The first is an LSTM network which models the user's intent. The second one is a belief tracker which is maintaining a multinomial distribution over a given value set. The policy network will aggregate the intent value, belief value, and the database result to generate the eventual output.

In \cite{williams2016end}, William et al propose a model for the task-oriented system which is using an LSTM to automatically predict the state of the system which doesn't require explicit representation or extra belief tracker. The LSTM can be optimized by supervised learning which requires the expert to provide the high-quality training data. Or it can guide the reinforcement learning system which just needs the users' input which could be accelerated by supervised learning(SL). The author also suggests this system is ready to active learning strategies which can boost the performance prominently.

In \cite{li2017end}, the author proposed an LSTM-based system which using the DQN framework to control the noise introduced by the user. The author remarks that this model's dialogue management module can directly interact with the database. The overall error can also be reduced due to the system structure.

In \cite{sukhbaatar2015end}, the author introduces a neural network model with recurrent attention model over a large external memory. It is trained end-to-end which is different from original Memory Netowrk[Memory]. The question sentence is represented as a bag of words and then converted to embeddings. The system uses a stacked structure to predict the answer to the eventual result. This system can achieve a comparable result in the QA tasks as LSTM or RNN models.

In \cite{bordes2016learning}, the author uses the memory network for QA into the dialog. The author uses a stacked structure of the memory matrix which is similar to \cite{sukhbaatar2015end}. The history of the dialogue is stored by the network. This is not a generative model and the result is selected from a given set of answers. The result of this method is still not perfect which is indicated by the authors.

In \cite{dodge2015evaluating}, the author proposed a new benchmark to evaluate the performance of the end-to-end dialogue systems and tested a series models on this new dataset. This new dataset consists of a variety of scenario including factual questions from movie database, personalization questions from the movieLens, short conversations, and natural dialogue from the Reddit. This dataset covers 75k movie entities and contains 3.5M training examples. Based on the author's test, the Memory Network[memory] has the best overall performance.
\subsubsection{External database and knowledge bases}
In order to construct a task-oriented dialogue system, the external database access is often necessary. Traditionally, researchers will use the symbolic representation to convert the sentences into database queries. With the fast growth of the neural-based dialogue systems, the neural model is also widely tested in the effort of integrating external databases and knowledge bases into dialogue systems. Various types of knowledge bases and databases have been explored.
\cite{lowe2015incorporating,ghazvininejad2017knowledge, han2015exploiting, zhu2017flexible, young2017augmenting,vougiouklis2016neural, yin2015neural, yin2015neural1, dhingra2016end, eric2017key}. 
To obtain a high-quality database, pre-processing steps such as data cleaning \cite{wang2016cleanix,wang2014cleanix,wu2019improving}, entity extraction \cite{wang2019efficient,wang2020boosting,wu2019scalable}, abnormal value detection \cite{li2019mining} with scalable distributed platforms \cite{li2018rios,das2019bigdata,wang2021formal,zaniolo2019monotonic,li2021kddlog,gulzar2017automated} are necessary.
The different format of knowledge bases include the key-value pairs and unstructured text corpus etc.

In \cite{lowe2015incorporating}, Lowe et al present a method to utilize the external unstructured text information to improve the neural-based dialogue system. The original system is an extension of a dual-encoder model which encode the context and the response by using two RNN networks. The extended version constructs one RNN component to encode the external knowledge. The relevance of external sentence is decided by a module which combines the hashing and TF-IDF techniques.

In \cite{ghazvininejad2017knowledge}, the author generalizes a seq2seq model by conditioning on both conversation history and external facts to build a chatbot which is fit for the open-domain settings. The external data is a large number of raw text entries from Foursquare, Wikipedia or Amazon reviews. These data are indexed by the named entity as the keys. Each conversation history will be processed to find the most important entity and a retrieve-based method is adopted to find the most relevant facts. The most relevant facts and conversation history will be encoded by the neural architecture.

In \cite{han2015exploiting}, Freebase is used as an external knowledge base. The named entity recognition will be applied to identify the key entity in a sentence and then the relevant entity will be extracted from the knowledge base which enables the system to answer more specialized questions.

In \cite{zhu2017flexible}, the author proposed a generative dialogue system which can handle the entities which do not appear in the training dataset. This work also adopts an entity extraction techniques but a more complex external entity selection function is designed so that it is more robust to the unseen entity. This method is also good for the serendipities which tend to punish similar results.

In \cite{young2017augmenting}, the author is trying to augment the chit-chat dialogue by using the common sense knowledge.

\subsection{Fine-aspects models in Conversational AI}
\subsubsection{Dialogue State Tracking}
The dialogue state tracking(DST)[42]\cite{henderson2015machine} is well adopted in various dialogue systems to track the state of ongoing conversations. It is a well-studied topic in dialogue processing. 
The dialogue state tracking is implemented by various techniques. With the booming of the neural network, RNN is widely adopted to model the internal belief of the dialogue. The dialogue state tracking has been formalized as a sequential annotation problem.

In \cite{henderson2014word}, Henderson et al propose a word-based tracking method which maps directly from dialogue state to the dialogue state without using any decoding steps. This method is based on an RNN structure which is able to generalize to unseen dialogue state. This method is able to achieve consistently good result among different metrics.

In \cite{mrkvsic2015multi}, the author proposed method for training multi-domain RNN dialogue state tracking models. This procedure will first use a complete dataset which contains all types of data to train a general model. Then the model will be trained by the domain-specific dataset to gain the specialized features while still keep the general cross-domain features. The experiment shows these models have a robust performance across all domains. They usually outperform the models that trained on target-domain alone.

In \cite{lee2016task}, the author proposed a state lineage tracking method in which the state of a dialogue is represented as a dynamic growing tree structure. This statistical solution enables the model to process more complex goal.

\subsubsection{Fine-grained response model}
RL-based dialog systems and end2end dialog systems have a common shortage. They oversimplify the dialog in our daily life. A lot of research is aiming at improving the generated dialog quality by introducing various constraints which are hinted by real word conversation's intuition and experiences. Such aspects include emotion, persona, diversity or conversation's context.[reference].

\textbf{Context}
A conversation is an alternating speaking process and the information is flowing in between the two dialog parties. Context is usually referred as the information which had been mentioned previously. Clearly, the context is very critical for response generation. Otherwise, the conversation tends to be off-topic or irrelevant at all. 

In \cite{sordoni2015neural}, the author implements a recurrent language model(RLM) which incorporate the context information to generate the new sentences. Three algorithms are presented in the paper. First one trains the RLM by using the concatenation of input, context, and output. In the generation phase, only input and context are feeding into the RLM to generate a proper hidden state of RLM and the output is generated by the hidden state. The second method maps the input and context from a bag of words to a fixed length vector by using a feed-forward neural network and update the hidden state by using such vectors. The third method maps input and context into two separate bags to generate vectors.

In \cite{duvsek2016context}, the author introduced an LSTM structure plus attention mechanism to encode the input and the context to generate desirable output sentences. One noticeable point of this work is using an n-gram reranker to generate the output which encourages output show overlap with the context. Another notable work is \cite{serban2017hierarchical}. In this paper, the author proposed a Latent Variable Hierarchical Encode Decoder Model which incorporates latent variables in the decoder.

\textbf{Persona}
The persona is another important dimension in conversational AI design. When we have a flat assumption of the persona of the AI, it tends to generate impersonal responses. For example, when a user tasks about frustrated experiences in browsing the Internet, a vanilla chatbot will reply "Thanks for your information" or "I have no idea what are you talking about". A robot trained with a specific persona will be likely to give a meaningful response, such as "Sorry to hear about it. It is usually caused by browser cache. Have you cleaned your browser cache yet?"

In \cite{luan2016lstm} Luan et al build a system with two different roles, the questioner, and the requester to train the RNN model. Topic modeling is introduced to represent the context of dialogue. In [27]\cite{luan2017multi}, the author adopts a joint training strategy for two models which share the same parameters in the decoding layer. The first model is trained on the general conversation data and the second model is trained on the task-specific datasets which aim to reduce nonsensical general responses. In [28]\cite{li2016persona}, Li proposed a persona-based model which aims to improve the consistency of the conversation. In this paper, the author proposed a speaker model and a speaker-addressee model to maintain the consistency of the conversation.

\textbf{Diversity}
The diversity is also an important aspect of the conversation. A conversation could be very boring if the response is too generic or universal. For example, if a chatbot replies "I don't know" or "This is a good question", the quality of conversation is hindered. In [30]\cite{li2015diversity}, Li proposed a new objective function for the seq2seq model, A Maximum Mutual Information(MMI) to replace the traditional log-likelihood function which often yields to generic responses. 

Zhao in \cite{zhao2017learning} proposed a conditional variational autoencoder model(CVAE) to generate more informative responses. He is attempting to model the conversation process as a one-to-many problem in the discourse level.

In \cite{mou2016sequence}, Mou proposed an interesting method which generates a most relevant noun about the question. The answer will be generated based on this noun.

\textbf{Emotion}
The emotion is a very important aspect of human-computer interface design. It is also true for conversational AI which is a prominent type of interface in between human and machine. The emotion could be expressed by a certain response style and word selection. 

In \cite{ghosh2017affect}, Ghosh modeled the affection by using an LSTM model which is conditioned on several predefined affection categories. The affection category is inferred by the word in the question. There is also an intensity variable to control the strength of the emotional words. 

In \cite{zhou2017emotional}, the author proposed an algorithm to generate the response based on predefined categories. Three aspects are considered to generate a proper emotional response. First one is the high-level abstraction of emotion expression which is modeled by embedding emotion categories. The second is capturing the internal emotional states and the last one uses explicit emotion expression with external emotion vocabulary. 

In \cite{asghar2018affective}, Asghar utilizes a hand-crafted affective dictionary which maps over 10,000 English words into a 3D space of valence, arousal, and dominance. Several different loss functions are attempted to optimize the performance such as minimizing affection dissonance, maximizing affection dissonance, maximizing affective content, affectively diverse decoding and diverse beam search.

%% file: datasets.tex
\section{Existing Datasets and benchmarks}

\subsection{Common datasets}
OpenSubtitles\cite{Tiedemann:RANLP5} is a public dataset which includes the subtitles for movies. It includes 59 kinds of languages for more than ten thousands movies. 

Ubuntu Dialogue Corpus, mentioned in \cite{lowe2017training}, includes conversations extracted from chats about Ubuntu technical support. The dataset can be accessed and generated following the instructions \cite{instruction} easily. The conversation is multi-turn and can involve multiple participants. There are about 936,000 dialogues and 100,000,000 contained in the dataset \cite{lowe2017training}. The least turn for each dialogue is 3 while the average is 7.71 as the stats information in Table 1 of \cite{lowe2017training}.   

\subsection{Open competitions}

Chatbot development challenge, hosted by industrial and academia, such as Amazon Alexa Prize, NIPS Conversational Intelligence Challenge and DSTC6 Dialog Systems Technology Challenge are prevalent.

Alexa Prize is held by Amazon since 2017. The challenge is building a chatbot which can conduct a conversation with human being on popular social topics for more than 20 minutes.\footnotetext{https://developer.amazon.com/alexaprize/}

\section{Concluding remarks}
In this paper, we briefly conclude the current progress of the conversational AI. Deep learning techniques such as sequence-to-sequence model and reinforcement learning are widely adopted in both general-purpose and goal-oriented conversational AI. With the adoption of the distributed representation, the internal state tracking problem has found a new promising direction. The incorporation of external information is still a big challenge. How to incorporate the external information with the neural-based conversational AI will be a valuable question.